\pdfoutput=1

\documentclass[11pt]{article}

\usepackage{emnlp2023}

\usepackage{times}
\usepackage{latexsym}

\usepackage[T1]{fontenc}

\usepackage[utf8]{inputenc}

\usepackage{microtype}

\usepackage{inconsolata}

\usepackage{natbib}
\usepackage{amsmath,amssymb,amsfonts}
\usepackage{textcomp}
\usepackage{graphicx}
\usepackage{xcolor}
\usepackage{algorithmic}

\usepackage{cite}
\usepackage{hyperref}

\usepackage{multirow}
\usepackage{tabularx}
\usepackage{booktabs}
\usepackage{xspace}
\usepackage{csvsimple}
\usepackage{enumitem}

\usepackage{cleveref}
\crefformat{section}{#2Sec.~#1#3}
\Crefformat{section}{#2Sec.~#1#3}
\crefformat{figure}{#2Fig.~#1#3}
\crefname{figure}{Fig.}{Figs.}
\Crefformat{figure}{#2Fig.~#1#3}
\crefformat{table}{#2Tab.~#1#3}
\Crefformat{table}{#2Tab.~#1#3}
\crefformat{appendix}{#2Appx.~#1#3}
\Crefformat{appendix}{#2Appx.~#1#3}

\setlist{nosep,leftmargin=*}

\newcommand{\name}{Antarlekhaka\xspace}

\newcommand{\comment}[1]{}

\newcommand{\dcsramayanaverses}{18754\xspace}

\newcommand{\dcsramayanachapters}{606\xspace}

\newcommand{\vrtasktypes}{5\xspace}

\newcommand{\vrannocount}{119\xspace}
\newcommand{\vrannosb}{1972\xspace}
\newcommand{\vrannosbverses}{1499\xspace}
\newcommand{\vrannocw}{1886\xspace}

\newcommand{\vrontoner}{89\xspace}
\newcommand{\vrannoner}{1717\xspace}
\newcommand{\vrannonerverses}{947\xspace}
\newcommand{\vrannocoref}{2271\xspace}
\newcommand{\vrannocorefverses}{950\xspace}

\newcommand{\vrannoagrels}{720\xspace}
\newcommand{\vrannoags}{90\xspace}
\newcommand{\vrannoagverses}{71\xspace}

\graphicspath{{graphics/}}

\title{Antarlekhaka: A Comprehensive Tool for\\ Multi-task Natural Language Annotation}

\author{Hrishikesh Terdalkar \and Arnab Bhattacharya \\
\texttt{\{hrishirt, arnabb\} @cse.iitk.ac.in} \\
  Department of Computer Science and Engineering\\
  Indian Institute of Technology Kanpur\\
  Kanpur, India
}

\begin{document}

\maketitle

\begin{abstract}
One of the primary obstacles in the advancement of Natural Language
	Processing (NLP) technologies for low-resource languages is the lack of
	annotated datasets for training and testing machine learning models. In this
	paper, we present \emph{Antarlekhaka}, a tool for manual annotation of a
	comprehensive set of tasks relevant to NLP. The tool is Unicode-compatible,
	language-agnostic, Web-deployable and supports distributed annotation by
	multiple simultaneous annotators. The system sports user-friendly interfaces
	for 8 categories of annotation tasks. These, in turn, enable the annotation
	of a considerably larger set of NLP tasks. The task categories include two
	linguistic tasks not handled by any other tool, namely, sentence boundary
	detection and deciding canonical word order, which are important tasks for
	text that is in the form of poetry. We propose the idea of \emph{sequential
	annotation} based on small text units, where an annotator performs several
	tasks related to a single text unit before proceeding to the next unit. The
	research applications of the proposed mode of multi-task annotation are also
	discussed. Antarlekhaka outperforms other annotation tools in objective
	evaluation.	It has been also used for two real-life annotation tasks on two
	different languages, namely, Sanskrit and Bengali. The tool is available at
	\url{https://github.com/Antarlekhaka/code}.
\end{abstract}
 
\section{Introduction and Motivation}\label{sec:introduction}

Manual annotation plays an important role in natural language processing (NLP).
It is particularly important in the context of low-resource languages for the
creation of datasets.

There are a number of syntactic and semantic tasks in NLP
which can utilize annotation by domain experts. Lemmatization, morphological
analysis, parts-of-speech tagging, named entity recognition, dependency parsing,
constituency parsing, co-reference resolution, sentiment detection, discourse
analysis and so on are some examples of such common NLP tasks.

There's a need of considering historical context and respecting the perspectives
of Indigenous language speaking communities when conducing NLP research
involving these languages \citep{schwartz2022primum}. Sanskrit, a classical
language, has a large amount of text available digitally; however, it still
suffers from poor performance in standard NLP tasks. Hence, manual annotation of
text in Sanskrit is of prime necessity. Further, most of the classical Sanskrit
literature is in poetry form following mostly free word order
\citep{kulkarni2015free}, without any punctuation marks. Therefore, certain
specialized tasks, such as sentence boundary detection and canonical word
ordering, are needed for Sanskrit text processing

\begin{figure*}[t]
    \centering
    \includegraphics[width=\linewidth]{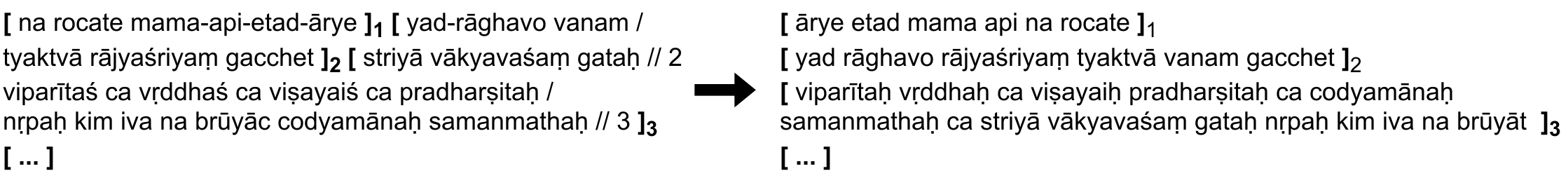}
    \caption{Sanskrit verses from Valmiki Ramayana. Original text appears on the
    left with sentence boundary markers added. The canonical word order is shown
    on the right.}
    \label{fig:ramayanaexample}
    \vspace*{-3mm}
\end{figure*}

Consider an example\footnote{Using IAST transliteration scheme:
\url{https://en.wikipedia.org/wiki/International\_Alphabet\_of\_Sanskrit\_Transliteration}}
from a Sanskrit text, \emph{Valmiki Ramayana}, \citep{dutt1891ramayana} shown in
\cref{fig:ramayanaexample}. The sentence boundaries are denoted using square
brackets ([ and ]). The half-verse boundaries are marked by single forward
slashes (/) and the verse boundaries by two forward slashes (//). It can be
observed that the sentence boundaries do not coincide with the verse boundaries.
In particular, there may be multiple sentences present in a single verse, or a
sentence may extend across multiple verses. Further, the right side of the arrow
shows that the canonical word order is different from the order in which words
appear in the original text.

For such languages that either do not use punctuations or use them in a limited
amount, sentence boundary detection is an important task. Additionally, in
languages with relatively free word order, decision of a canonical word
order is also relevant. These two tasks also play a vital role when dealing with
the corpora in the form of poetry, making them potentially relevant for all
languages.

Performing multiple annotation tasks on the same corpus is common, and the order
of these tasks can be important due to their interdependence. Specifically, in
cases\footnote{For corpora without clear sentence boundaries, like languages
with limited punctuation or poetic corpora} where sentence boundary detection is
relevant, it must be performed before any other annotation task. For instance,
determining the word order of a sentence requires finalizing the constituent
words first. The same holds true for tasks such as dependency parsing, sentence
classification, and discourse analysis.

\begin{table*}[t]
    \centering
    \caption{Comparison of NLP annotation tools based on primary features and
    supported tasks}
    \label{table:comparison}
    \vspace*{-2mm}
\resizebox{0.70\linewidth}{!}{
    \begin{tabular}{lccccccc}
        \toprule
        ~ & INCEpTION & GATE & BRAT & FLAT & doccano & Sangrahaka & \bf \name\\
        \midrule
        Distributed Annotation & \checkmark & \checkmark & \checkmark & \checkmark & \checkmark & \checkmark & \checkmark\\
        Easy Installation & ~ & ~ & \checkmark & \checkmark & \checkmark & \checkmark & \checkmark\\
        Sequential Annotation & ~ & ~ & ~ & ~ & ~ & ~ & \checkmark\\
        Querying Interface & ~ & ~ & ~ & ~ & ~ & \checkmark & ~\\
        \midrule
        Token Text Annotation & \checkmark & \checkmark & \checkmark & \checkmark & ~ & ~ & \checkmark\\
        Token Classification & \checkmark & \checkmark & \checkmark & \checkmark & \checkmark & ~ & \checkmark\\
        Token Graph & \checkmark & \checkmark & \checkmark & \checkmark & ~ & \checkmark & \checkmark\\
        Token Connection & \checkmark & \checkmark & \checkmark & \checkmark & ~ & \checkmark & \checkmark\\
        Sentence Boundary & \checkmark & ~ & ~ & ~ & ~ & ~ & \checkmark\\
        Word Order & ~ & ~ & ~ & ~ & ~ & ~ & \checkmark\\
        Sentence Classification & \checkmark & ~ & ~ & ~ & ~ & ~ & \checkmark\\
        Sentence Graph & ~ & ~ & ~ & ~ & ~ & ~ & \checkmark\\
        \bottomrule
    \end{tabular}
    }
\vspace*{-2mm}
\end{table*}

In this paper, we describe \emph{\name}\footnote{\emph{\name} is a Sanskrit word
meaning `annotator'.}, a tool for distributed annotation that provides
user-friendly interfaces to facilitate the annotation process of various common
NLP tasks in a straightforward and efficient way. We propose a sequential
annotation model, where an annotator carries out multiple annotation tasks
relevant for a small text unit, such as a verse, before proceeding to the next.
The tool has full Unicode support and is designed to be language-agnostic,
meaning it can be used with corpora from any language, making it highly
versatile. The tool sports eight task-specific user-friendly annotation
interfaces corresponding to eight general categories of NLP tasks: sentence
boundary detection, canonical word ordering, free-form text annotation of
tokens, token classification, token graph construction, token connection,
sentence classification and sentence graph construction. The goal of the tool is
to streamline the annotation process, making it easier and more efficient for
annotators to complete multiple NLP tasks on the same corpus. Annotators can
participate in the annotation without any programming knowledge. Additionally,
the tool's easy setup and intuitive administrator interface make it accessible
to administrators with minimal technical expertise.
 
\section{Background}\label{sec:related}

An annotation tool is crucial for the successful completion of any annotation
task, and its success relies heavily on its usability for the annotators. Apart
from this, the tool should be easily installable and should support web
deployment for distributed annotation, allowing multiple annotators to work on
the same task from different locations. The administration interface of the tool
should also be intuitive and should provide convenient access to common
administrative tasks such as corpus upload, ontology creation, and user access
management. Additionally, often there is a need for multiple annotation tasks to
be completed on the same corpus.
A well-designed annotation tool should encompass these features to ensure a
smooth, efficient, and accurate annotation process.

Numerous text annotation tools are available that target specific annotation
tasks, such as WebAnno \citep{yimam2013webanno}, INCePTION
\citep{klie2018inception}, GATE Teamware \citep{bontcheva2013gate}, FLAT
\citep{flat2014}, BRAT \citep{stenetorp2012brat}, doccano \citep{doccano} and
more. However, each of these tools falls short in fulfilling all the
requirements of an ideal annotation tool. For instance, WebAnno is rich in
features but becomes complex to use and experiences performance issues as the
number of lines displayed on the screen increases. Both WebAnno and BRAT lack
full support for Firefox \citep{fort2016collaborative}, an issue that was
rectified in INCEpTION. GATE Teamware suffers from shortcomings such as
inadequate support for relation and co-reference annotation
\citep{herik2018agents}, installation issues \citep{neves2021extensive} and
complexity for average users \citep{yimam2013webanno}. FLAT uses a non-standard
FoLiA XML data format and the system is not intuitive
\citep{neves2021extensive}. BRAT has not been actively\footnote{The latest
version was published  in 2012} developed and exhibits issues such as slowness,
limited scope for configuration and limitations regarding file formats
\citep{yimam2013webanno}. The tool doccano, although simple to set up and
intuitive, only supports labeling tasks. Sangrahaka
\citep{terdalkar2021sangrahaka}, while being easy to set up and use, focuses
only on the annotation towards creation of knowledge graphs and lacks support
towards general-purpose NLP annotation tasks.

Some annotation tools including INCEpTION use spans for marking most
annotations, which a user by selecting and dragging mouse cursor over the corpus
text. This method, while versatile, has a trade-off that the annotation process
is slower and more tedious. Importantly, none of these tools address crucial
tasks like canonical word ordering. Hence, there is a need for an annotation
tool that is user-friendly, easy to install and deploy, and encompasses all the
necessary tasks for NLP annotation.

Thus, for the general purpose multi-task annotation of NLP tasks, we present
\emph{\name}. The annotation is performed in a sequential manner on small units
of text (e.g., verses in poetry). The application is language and corpus
agnostic. The tool is able to process data in two different formats: the
standard CoNLL-U\footnote{\url{https://universaldependencies.org/format.html}}
format and plain text format. Regular-expressions based tokenizer is applied
when using the data in plain text format.

\cref{table:comparison} shows a comparison of the prominent annotation tools. We
also conduct an objective evaluation of \emph{\name} using the scoring
methodology proposed by \citep{neves2021extensive}. We modify the criteria
suitable to the domain of NLP annotation. Details of the evaluation are
described in \cref{sec:evaluation}. It is important to note that while some of
the existing tools, in theory, have the capability to support certain NLP tasks,
they may not be designed with user-friendly interfaces.

\section{Architecture}\label{sec:system}

\emph{\name} is a language-agnostic, multi-task, distributed annotation tool
presented as a Web-deployable software. The tool makes use of several
technologies, including \emph{Python 3.8} \comment{\citep{python3}}, \emph{Flask
2.2.2}\comment{\footnote{\url{https://flask.palletsprojects.com/en/2.2.x/}}}
\comment{\citep{ronacher2011flask,grinberg2018flask}}, and \emph{SQLite 3.38.3}
\comment{\citep{sqlite2022hipp}} for the backend, and \emph{HTML5},
\emph{JavaScript}, and \emph{Bootstrap 4.6} \comment{\citep{bootstrap46}} for
the frontend.

\emph{Flask} web framework powers the backend of \emph{\name} providing a robust
and scalable infrastructure. A web framework is responsible for a range of
backend tasks, including routing, templating, managing user sessions, connecting
to databases and others. The recommended way to run the tool in a production
environment is using a \emph{Web Server Gateway Interface} (WSGI) HTTP server,
such as \emph{Gunicorn}\comment{\footnote{\url{https://gunicorn.org/}}}
\comment{\citep{gunicorn}}, which can operate behind a reverse proxy server such
as \emph{NGINX}\comment{\footnote{\url{https://nginx.org/en/}}} \comment{\citep{nginx}} or
\emph{Apache HTTP Server}\comment{\footnote{\url{https://httpd.apache.org/}}}
\comment{\citep{apache}}. However, any WSGI server, including the built-in
server of Flask, can be utilized to run the application.

\emph{SQLite} is used as the database management system to store and manage the
data and metadata associated with the annotation tasks. An Object Relational
Mapper (ORM) \emph{SQLAlchemy}\footnote{\url{https://www.sqlalchemy.org/}}
\comment{\citep{sqlalchemy}} is used to interact with the relational database.
This allows the user to choose any supported dialect of traditional SQL, such as
\emph{SQLite}, \emph{MySQL} \comment{\citep{mysql}}, \emph{PostgreSQL}
\comment{\citep{postgressql}}, \emph{Oracle} \comment{\citep{oracle}}, \emph{MS-SQL}
\comment{\citep{mssql}}, \emph{Firebird} \comment{\citep{firebird}},
\emph{Sybase} \comment{\citep{sybase}} and
others\footnote{\url{https://docs.sqlalchemy.org/en/20/dialects/}}.

The frontend of the tool, built using \emph{HTML5}, \emph{JavaScript}, and
\emph{Bootstrap}, provides user-friendly interfaces for annotators and
administrators. The tool provides a feature-rich administrative interface to
manage user access, corpus, tasks and ontology. The tool also includes eight
types of intuitive annotation interfaces, explained in detail in
\cref{sec:interfaces}.

The tool simplifies setup with a single configuration file that controls various
customizable aspects. Overall, by combining the state-of-the-art technologies,
\emph{\name} offers a powerful and flexible solution for large-scale annotation
projects.

\subsection{Workflow}

\begin{figure}[t]
    \centering
    \includegraphics[width=0.9\columnwidth]{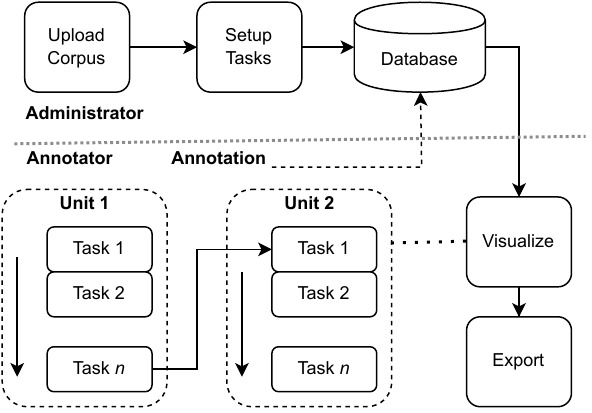}
    \caption{Workflow of \emph{\name}}
    \label{fig:architecture}
\vspace*{-4mm}
\end{figure}

The workflow of the system is demonstrated in \cref{fig:architecture}.
The application is presented as a full-stack web-based software. It follows a
single-file configuration system. An administrator may configure the tool and
deploy it to web, making it immediately available for use. User registration is
supported. User access is controlled by a 4-tier permission system, namely User,
Annotator, Curator and Admin.

The tool has eight annotation interface templates corresponding to eight generic
categories of NLP annotation tasks: sentence boundary detection, canonical word
order, free-form token annotation, token classification, token graph
construction, token connection, sentence classification, and sentence graph
construction. Various NLP tasks can be modelled using each of these categories.
More than one tasks of same category may be required for a specific annotation
project. For example, named entity recognition (NER) and parts-of-speech (POS)
tagging are both examples of token classification. To facilitate this, the
administrative interface allows an administrator to create multiple tasks of
each category. Additionally, an administrator can also control the set of active
tasks, order of the tasks, ontology for the relevant tasks, corpus management
and user access management.

We propose a streamlined sequential mode of annotation where an annotator
completes multiple tasks for a single unit of text before moving on
to the next unit. The set and order of tasks is customizable through an
administrative interface. We consider a small logical block of text as a unit
for the annotation, e.g., a verse from the poetry corpus.

\subsection{Data}

The data for corpus can be in either of two formats: \emph{CoNLL-U} format or
\emph{plain text} format and can contain Unicode text. \emph{CoNLL-U} is a
widely used format for linguistic annotation, and it is based on the column
format for treebank data. The format is designed to store a variety of
linguistic annotations, including part-of-speech tags, lemmas, morphological
features, and dependencies between words in a sentence.
Data in \emph{CoNLL-U} format can be obtained from treebanks such as
Universal Dependencies \citep{de2021universal}, which is a project that aims to
develop cross-linguistically consistent treebank annotation. In addition, NLP
tools such as Stanza \citep{qi2020stanza} are capable of processing a general
corpus of text and producing data in \emph{CoNLL-U} format. \comment{, making it easier to
obtain data in this format.}

\comment{On the other hand,} Plain text data is processed using a regular-expression based
tokenizer, which is a process that splits the text into individual units of
meaning, such as verses, lines and tokens using patterns defined in the form of
regular expressions to identify the respective separators. The plain text
processor module is a pluggable component. An administrator may reimplement it
using any language specific features or tools as long as the data output by the
module meets the current format specifications.

After the data has been imported, it is organized in a five-level hierarchy
structure consisting of: Corpus, Chapter, Verse, Line, and Token. The
hierarchical structure of the data provides a clear and organized framework for
annotating and analyzing the data, making it easier to capture the relationships
between different elements of the data.

\subsection{Task Categories and Interfaces}
\label{sec:interfaces}

\begin{figure*}
    \centering
    \includegraphics[width=0.70\textwidth]{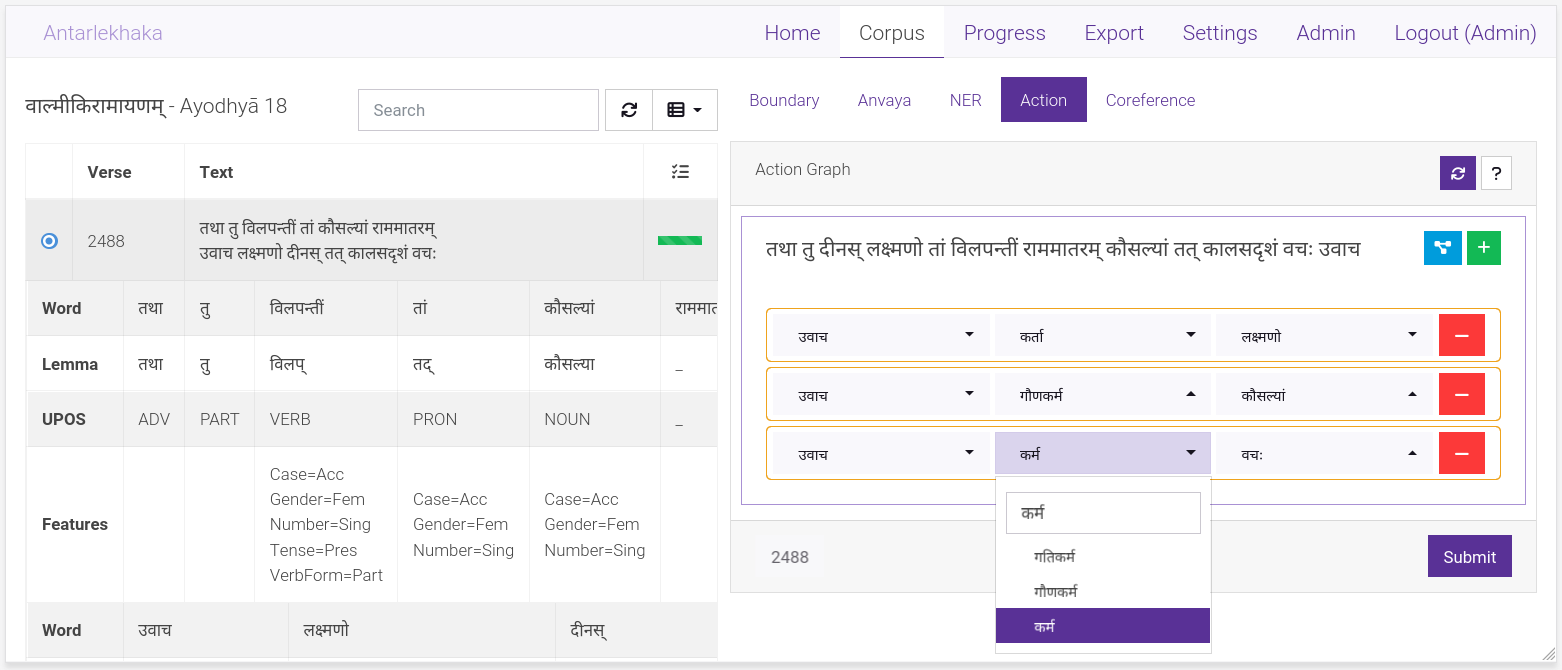}
    \caption{Annotation Interface: a Sanskrit corpus split into small units, and annotation area with task tabs}
    \label{fig:interface_annotation}
    \vspace*{-3mm}
\end{figure*}

The annotation supports annotation towards eight categories of annotation tasks
and offers intuitive interfaces for each category. Annotators view the corpus in
the form of text units (e.g., verses) on the left, and an annotation area on the
right. After submitting annotations for a task, the interface automatically
advances to the next task. Annotators are expected to complete all the tasks
associated with a text unit before moving on to the next unit. This, however, is
not strictly enforced, allowing annotator to still go back to modify
annotations. \cref{fig:interface_annotation} showcases the overall annotation
interface.

The administrator can configure task-related information, including task titles,
instructions, active tasks, and their order, through the administrative
interface. This interface is illustrated in
\cref{fig:interface_admin_manage_tasks} (\cref{sec:screenshots}). Tasks such as
user access management, corpus creation, and ontology management also have
intuitive administrative interfaces. Next, we provide a detailed description of
each task category and its corresponding interface.

\subsubsection{\textbf{Sentence Boundary Detection}}

The importance of the sentence boundary task is not limited to languages without
distinct sentence markers; it also pertains to poetry text, making it relevant
to all languages.

The annotator's task is to identify and mark sentence boundaries by placing the
delimiter `\textbf{\#\#}' (two `hash' symbols) at the end of each sentence in
the provided editable text area prefilled with the original text. If the
sentence does not end in the displayed unit, the user does not add any
delimiters. After marking sentence boundaries, the user can proceed to the next
annotation task. An illustration of this annotation task is shown in
\cref{fig:interface_boundary} (\cref{sec:screenshots}).

It is worth mentioning that although the sentence boundary task is given primary
citizen treatment, it can still be turned off for languages where it is not
applicable. In such instances, the boundaries of annotation text units (e.g.,
verses) are treated as sentence boundaries.

\subsubsection{\textbf{Canonical Word Order}}

All sentences that end in the current unit of text are displayed to the
annotator as a list of sortable tokens. The annotator can rearrange these tokens
into the correct canonical word order by dragging them into place. Additionally,
if any tokens are missing, the annotator can add them as well. A visual
representation of this task is shown in \cref{fig:interface_word_order}
(\cref{sec:screenshots}). The sorting capability is made possible through the
use of the \emph{jQuery UI (Sortable
plugin)}\footnote{\url{https://api.jqueryui.com/sortable/}}.

\subsubsection{\textbf{Token Annotation}}

The token annotation interface allows an annotator to add free-form text
associated with every token. This free-form text can have different purposes,
such as to identify the root word of a word (lemmatization), to separate
multi-word expressions into individual words (compound splitting), to analyze
the morphological structure of a word (morphological analysis), etc. The token
annotation interface is shown in \cref{fig:interface_token_annotation}
(\cref{sec:screenshots}).

\subsubsection{\textbf{Token Classification}}

Token classification is a process of assigning predefined categories to
individual tokens in text data. It is a special case of free-form token
annotation, wherein the annotations are guided by an ontology. For such a task,
an administrator must create an ontology.  During the annotation process, an
annotator is provided with a list of tokens, each accompanied by a dropdown
menu, from which they can select the appropriate category for relevant tokens.
Some common examples of token classification tasks include NER, dependency
tagging, POS tagging, and compound classification.
\cref{fig:interface_token_classification} (\cref{sec:screenshots}) illustrates
the token classification interface.

\vspace*{2mm}
\subsubsection{\textbf{Token Graph}}

A token graph is a graph representation of the sentence, where the nodes are
tokens belonging to a single sentence and the relations are based on an
ontology. Tasks such as dependency parse tree, constituency graph, action graph
are examples of tasks belonging to this category.

Semantic triple\footnote{\url{https://en.wikipedia.org/wiki/Semantic_triple}} is
a standard format to represent and store graph-structured information in a
relational database in a systematic manner. The interface allows an annotator to
add multiple relations per sentence in the form of subject-predicate-object
triples, where subject and object are tokens from the sentence and the predicate
is a relation from the task specific ontology. The valid values of subject,
object and predicate appear in individual dropdown menu elements for the
annotator to choose from. Erroneous triples may also be removed. During the
annotation process, an annotator can view the current status of the token graph
at any time using the `Show Graph' button.
\cref{fig:interface_token_graph} (\cref{sec:screenshots}) shows the token graph
interface with graph visualization.

\subsubsection{\textbf{Token Connection}}

Token connection is similar to token graph, however, there is a single type of
relation to be captured. For example, when marking co-references, only
connecting the two tokens to each other is sufficient, while the relationship
`is-coreference-of' is implicit. The tool provides a special simplified
interface for this scenario. In addition to implicit relations, token
connections can also be established across sentences. The annotator is presented
with a list of clickable tokens from the current sentence as well as tokens from
a context window of previous $n$ (a configuration parameter with default as $5$)
sentences. The annotator can add a connection by clicking on the source token
and the target token one after the other and confirming the connection. If a
connection is added in error, it can be removed as well. In some cases, a
connection might extend beyond the default context window. To address this, we
have incorporated a button that an annotator can click to load additional
context when needed. The token connection interface is shown in
\cref{fig:interface_token_connection} (\cref{sec:screenshots}).

\subsubsection{\textbf{Sentence Classification}}

Sentence classification is a task where sentences are classified into different
categories, e.g., sentiment classification and sarcasm detection. This task is similar to ontology-driven token classification, with
the difference being that classes are associated with sentences rather than
tokens. The ontology is predefined by the administrator while setting up the
task. The annotator can select the category for a sentence from a dropdown menu.
\cref{fig:interface_sentence_classification}
(\cref{sec:screenshots}) illustrates the sentence classification interface.

\subsubsection{\textbf{Sentence Graph}}

Sentence graph is a graph representation of relationships between sentences, captured as subject-predicate-object triples. The
connections can be between tokens or complete sentences.
Tokens from the previous $n$
(a configuration parameter with default as $5$) sentences are presented as
buttons arranged in the annotated word order. An annotator creates connections
by clicking on the source and target tokens and selecting the relationship from
a dropdown menu based on an ontology. A special token is provided to denote the
entire sentence as an object. Tasks such as timeline annotation and discourse
graphs are examples of tasks belonging to this category.
\cref{fig:interface_sentence_graph} (\cref{sec:screenshots}) shows the interface
for creating sentence graph connections. Similar to the token graph task, an
annotator can also visualize the sentence graph.
 
\subsection{Clone Annotations}
\label{sec:clone}

The administrative interface offers the capability to replicate annotations from
one user to another. This feature proves valuable in cases where certain
annotators possess expertise in specific tasks or if an annotator leaves a task
incomplete, requiring another annotator to resume the task from their account.
The cloned annotations are displayed just like regular annotations. However,
they maintain the source information, including the annotator's ID and a
reference to the original annotation.

\subsection{Pluggable Heuristics}
\label{sec:heuristics}

The tool supports the use of heuristics as `pre-annotations' to assist
annotators. Heuristics are custom functions that generate suggestions for the
annotators to use or ignore. These heuristics are often specific to the language
and corpus, and thus, must be implemented by the administrator when setting up
the tool. The codebase of the tool outlines the format and type specifications
of the heuristics, making them a pluggable component.

\subsection{Progress Report}
\label{sec:progress}

Detailed progress tracking serves multiple essential functions. Firstly, it
provides project managers with the capability to allocate resources effectively
and oversee the distribution of tasks among chapters. This facilitates the early
detection of potential bottlenecks or areas needing extra focus. Moreover, the
breakdown of progress contributes to streamlining the annotation process,
guaranteeing the steady advancement of all chapters and tasks at an optimal
rate. To facilitate this, we've developed an interface that offers a
comprehensive overview of annotators' progress. This interface provides a
detailed breakdown of advancements on a per-chapter and per-task basis, enabling
thorough tracking and evaluation of their contributions.

\subsection{Export}
\label{sec:export}

The export interface enables the access, retrieval and visualization of the
annotated data for each task in a clear and straightforward manner. Annotator
can easily view and export the data in two formats (1)~a human-readable format
for easy inspection and (2)~a machine-readable format compatible with the
standard NLP tools. The specifics of the standard format depend on the task. For
example, a standard format for NER datasets is the BIO format
\citep{tjong2003introduction}, which stands for \emph{begin}, \emph{inside}, and
\emph{outside}. The B-tag marks the beginning of a named entity, while the I-tag
indicates the continuation of a named entity. The O-tag signifies that a word is
not part of a named entity. \cref{fig:interface_export} (\cref{sec:screenshots})
illustrates the export interface showcasing the capability to export NER data in
the standard BIO format. The interface facilitates the visualization and export
of graph representations for tasks related to graphs.
The export interface is accessible not only to annotators but also to curators,
allowing them to review annotations made by other users. This feature
serves as a mechanism for quality control.

\begin{table*}[!h]
    \caption{Objective evaluation criteria for annotation tools. Each feature is
    evaluated on a ternary scale of 0, 0.5 and 1, where 0 indicates absence of
    the feature, 0.5 indicates partial support and 1 indicates full support for
    the feature.}
    \label{table:annoeval}
	\begin{center}
    \resizebox{0.80\textwidth}{!}{
    \begin{tabular}{llc|ccccc|c}
        \toprule
        \multicolumn{3}{c|}{\textbf{Criteria}} & \multicolumn{6}{c}{\textbf{Tools}}\\
        \cmidrule{1-3}\cmidrule{4-9}
        \textbf{ID} & \textbf{Description} & \textbf{Weight} & \textbf{INCEpTION} & \textbf{doccano} & \textbf{FLAT} & \textbf{BRAT} & \textbf{Sangrahaka} & \textbf{Antarlekhaka}\\
        \midrule
        \vspace*{-3mm}
        \begin{filecontents*}{comparison.csv}
ID,Description,Weight,INCEpTION,doccano,FLAT,BRAT,Sangrahaka,Antarlekhaka
P1,Year of the last publication,0,1,0,0,1,1,1
P2,Citations on Google Scholar,0,1,0,0,1,0,0
P3,Citations for Corpus Development,0,1,0,0,1,0,0
T1,Date of the last version,1,1,1,1,0.5,1,1
T2,Availability of the source code,1,1,1,1,1,1,1
T3,Online availability for use,1,0,0,1,0,0,0
T4,Easiness of Installation,1,0,1,1,0.5,1,1
T5,Quality of the documentation,1,1,1,1,1,0.5,0.5
T6,Type of license,1,1,1,1,1,1,1
T7,Free of charge,1,1,1,1,1,1,1
D1,Format of the schema,1,1,1,1,0.5,1,1
D2,Input format for documents,1,1,0.5,1,1,1,1
D3,Output format for annotations,1,1,1,1,0.5,0,0
F1,Allowance of multi-label annotations,1,1,0,1,1,1,1
F2,Allowance of document level annotations,0,0,0,0,0,0,0
F3,Support for annotation of relationships,1,1,0,0,1,1,1
F4,Support for ontologies and terminologies,1,1,0,1,1,1,1
F5,Support for pre-annotations,1,0.5,0,0.5,0.5,0,0
F6,Integration with PubMed,0,0,0,0,0,0,0
F7,Suitability for full texts,1,0.5,0.5,1,1,1,1
F8,Allowance for saving documents partially,1,1,1,1,1,1,1
F9,Ability to highlight parts of the text,1,1,1,1,1,1,1
F10,Support for users and teams,1,0.5,0.5,1,0.5,0.5,0.5
F11,Support for inter-annotator agreement,1,1,0.5,0,0.5,0.5,0.5
F12,Data privacy,1,1,1,1,1,1,1
F13,Support for various languages,1,1,1,1,1,1,1
A1,Support for querying,1,0,0,0,0,1,0
A2,Crash tolerance,1,0,0,0,0,1,0.5
A3,Web-based / Distributed annotation,1,1,1,1,1,1,1
A4,Sequential Annotation,1,0,0,0,0,1,1
A5,Support for Sentence Boundary Annotation,1,1,0,0,0,0,1
A6,Support for Word Order Annotation,1,0,0,0,0,0,1
A7,Support for Token Classification Tasks,1,1,1,1,1,1,1
A8,Support for Sentence Classification Tasks,1,1,0,0,0,0,1
\end{filecontents*}
\csvreader[head to column names]{comparison.csv}{}{\\\ID&\Description&\Weight&\INCEpTION&\doccano&\FLAT&\BRAT&\Sangrahaka&\Antarlekhaka}\\
        \midrule
        \multicolumn{2}{c}{\textbf{Total}} & $29$ & $21.5$ & $16.0$ & $20.5$ & $18.5$ & $21.5$ & $\mathbf{23.0}$\\
        \multicolumn{2}{c}{\textbf{Score}} & ~ & $0.74$ & $0.55$ & $0.71$ & $0.64$ & $0.74$ & $\mathbf{0.79}$\\
        \bottomrule
    \end{tabular}
    }
\end{center}
\vspace*{-3mm}
\end{table*}

\subsection{Language Independence}

Unicode is a widely used standard for encoding, representing, and handling text
in a uniform and consistent way across various languages, computing platforms
and applications. The standard assigns unique numerical codes to each character
in a large number of scripts.  Thus, full Unicode support allows users to work
with text data in their preferred languages.
 
\section{Evaluation}\label{sec:evaluation}

The tool is being used for annotation of a large corpus in Sanskrit, namely,
\emph{Valmiki Ramayana}. The details of this project are described in
\cref{sec:casestudy}. Additionally, the tool is also being used for the
annotation of plain text corpus in Bengali language.

Following \citep{terdalkar2021sangrahaka}, we have evaluated our tool using a
two-fold evaluation method of subjective and objective evaluation.  We have not
used the time taken for annotation as an evaluation metric since annotators
often spend more time processing the text to identify the relevant information
than physically annotating.  Hence, it may not be a reliable measure.

\subsection{Subjective Evaluation}

\begin{figure}
	\vspace*{-3mm}
	\centering
	\includegraphics[width=0.95\columnwidth]{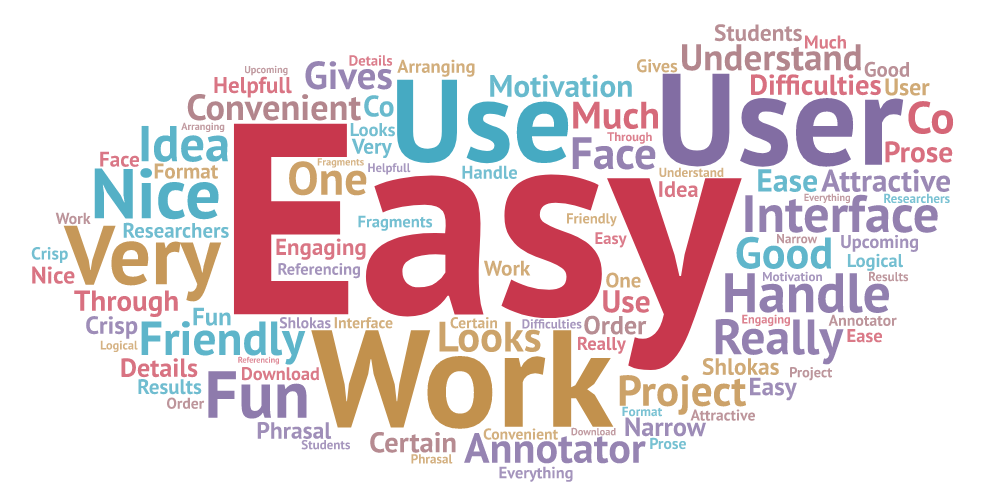}
	\caption{Word cloud of comments received in a survey}
	\label{fig:wordcloud}
	\vspace*{-5mm}
\end{figure}

For the subjective evaluation, an online survey was conducted, with
participation from $16$ annotators. They were asked to rate the tool on a scale
of 1 to 5 across various categories including ease of use, annotation interface,
and overall performance. The feedback from annotators was predominantly
positive. The tool received a score of $4.3$ for ease of use, $4.4$ for
annotation interface, and an overall score of $4.1$. Furthermore, we gathered
comments from users, and a word cloud over these comments can be found in
\cref{fig:wordcloud}.

\subsection{Objective Evaluation}

The objective evaluation utilized a scoring mechanism similar to that employed
in previous studies \citep{neves2021extensive,terdalkar2021sangrahaka}. We
retained the additional categories introduced by \citep{terdalkar2021sangrahaka}
while incorporating supplementary categories pertinent to the comprehensive
assessment of a general-purpose NLP tool. These supplementary categories were
designed to gauge the tool's support for a diverse range of NLP tasks.
Consequently, a total of $29$ categories were employed for the evaluation
process. Each tool was assigned a rating: 1 for full support of a feature, 0.5
for partial support, and 0 for the absence of a feature.

In this evaluation, \emph{\name} achieved a score of $0.79$, outperforming other
tools like \emph{Sangrahaka} ($0.74$), \emph{INCEpTION} ($0.74$) and \emph{FLAT}
($0.71$). The detailed list of the 29 categories used for objectively scoring
the annotation tools can be found in \cref{table:annoeval}.

\subsection{Case Study: Large Sanskrit Text}
\label{sec:casestudy}

The \emph{Valmiki Ramayana}, an ancient Sanskrit text, presents a rich and
intricate narrative with a diverse array of characters, events, emotions, and
settings, making it an ideal choice for annotation purposes. We employ \name,
for this large-scale annotation task, as a case study. The text, sourced from
the Digital Corpus of Sanskrit (DCS) \citep{hellwig2010dcs}, comprises a total of
\dcsramayanaverses verses distributed across \dcsramayanachapters chapters.

The annotation \comment{of the \emph{Valmiki Ramayana}} is being performed with the help
of \vrannocount annotators to annotate \vrtasktypes tasks per verse. These
annotations have resulted in the following \emph{task-specific datasets}:

\begin{itemize}
	\item \emph{Sentence Boundary}: \vrannosb sentence markers across
		\vrannosbverses verses;
	\item \emph{Canonical Word Ordering}: \vrannocw sentences;
	\item \emph{Named Entity Recognition}: \vrannoner entities in
		\vrannonerverses verses based on custom ontology with \vrontoner labels;
	\item \emph{Co-reference Resolution}: \vrannocoref co-reference connections
		across \vrannocorefverses verses;
	\item \emph{Action Graph}: \vrannoags action graphs with \vrannoagrels
		relations from \vrannoagverses verses, where an \emph{action graph}
		captures action-related words, encompassing verbs, participles, and
		other action-denoting words and their relationships with other words.
\end{itemize}

We continue to enhance these datasets through the ongoing annotation endeavour.

\vspace*{-1mm}

\section{Potential for NLP Research}
\label{sec:research}

A natural language annotation tool simplifies the process of creating datasets
for machine learning models, which is useful for NLP tasks such as
lemmatization, NER, POS tagging, co-reference resolution, text classification,
sentence classification, and relation extraction. Annotator-friendly and
intuitive interfaces can simplify the otherwise tedious process of manual
annotation process to a great extent. \comment{High-quality, manually annotated training
datasets contribute directly towards improving the accuracy of NLP models.}
The effectiveness of higher-level tasks such as question-answering,
grammatical error correction and machine translation often relies on
the success of several low-level tasks, which can be handled by a multiple task
annotation tool. \comment{For example, domain-specific QA often involves building
knowledge graphs \citep{voorhees1999trec, hirschman2001natural,
kiyota2002dialog, yih2015semantic}, which requires performing NER, co-reference
linking, POS tagging, and dependency relation identification on the same corpus.}
The tool's ability to handle large amounts of data and multiple users
simultaneously can contribute to faster completion of these tasks.

\vspace*{-1mm}

\section{Conclusions}
\label{sec:conclusion}

\vspace*{-1mm}

We have developed a web-based multi-task annotation tool called \emph{\name} for
sequential annotation of various NLP tasks. It is available at
\url{https://github.com/Antarlekhaka/code}. The tool is language-agnostic and
has full Unicode support. The tool sports eight categories of annotation tasks
and an annotator-friendly interface for each category. Multiple annotation tasks
from each category are supported. The tool enables creation of datasets for
computational linguistics tasks without expecting any programming knowledge from
the annotators or administrators. It is actively being used for a large-scale
annotation project, involving a large Sanskrit corpus and a significant number
of annotators, as well as another annotation task in Bengali language.
\emph{\name} has a potential to propel research opportunities in NLP by
simplifying the conduction of large-scale annotation projects.

\comment{

The tool is actively being used for a large-scale annotation project, involving
a large Sanskrit corpus and a significant number of annotators, as well as
another annotation task in Bengali language. The tool is also actively
maintained. In the future, we plan to integrate various state-of-the-art NLP
tools to add out-of-the-box support for several languages and aid annotators by
providing suggestions.

}

\section{Limitations}
\label{sec:limitations}

While \emph{\name} is a powerful tool for annotation, it does have some
limitations. These include:

\begin{itemize}
\item \emph{Subjectivity in annotations}: Manual annotation can introduce
    subjective biases and inconsistencies among annotators. Currently,
    there is no in-built automated mechanism to ensure inter-annotator agreement
    and quality check is performed manually with a small set of curators.
    \item \emph{Language-specific challenges}: \name may encounter challenges specific to certain languages
    or linguistic phenomena, including variations in syntax,
    morphology, or semantic nuances that may require additional customization or
    fine-tuning. Additionally,
    languages with complex orthographic systems or unique writing conventions
    may pose difficulties.
    \item \emph{Dependency on sequential annotation}: Sequential annotation is
    one of the strong features of the tool. However, if the sentence boundary
    detection task is enabled, it imposes a dependency in the sense that
    other tasks can be performed only after marking the sentence boundaries.
\end{itemize}
 
\section{Ethics Statement}

The research conducted in the development and use of \emph{\name} adheres
to ethical considerations and guidelines. The annotation tasks performed using
the tool involve the analysis and processing of language data. We ensure the
following ethical principles:

\begin{itemize}
    \item \emph{Informed Consent}: Prior to engaging in annotation tasks, all
    annotators participating in the research are informed about the nature of
    the tasks, their purpose, and the potential use of the annotated data.
    Annotators provide their voluntary consent to participate.
    \item \emph{Anonymity and Privacy}: All personal information of annotators is kept
    confidential and handled securely. The data collected during the annotation
    process is anonymized to protect the privacy of individuals involved.
    \item \emph{Data Usage}: The annotated data is solely used for research purposes
    and in compliance with relevant data protection regulations. It is not
    shared with any third parties without explicit consent or legal
    requirements.
    \item \emph{Bias Mitigation}: We strive to minimize any biases that may
    arise during the annotation process. Annotators are provided with guidelines
    and training to ensure consistency and fairness in their annotations.
    Regular quality checks are being performed to address any potential bias issues for the Sanskrit text corpus.
\item \emph{Annotator Pool}: Annotators for the Sanskrit text corpus are under-graduate and post-graduate level Sanskrit students from various institutes and colleges. This ensured that annotations were of accepted quality. Participation in the annotation task was voluntary, and everybody who wanted to annotate was allowed to do so.
\end{itemize}

By adhering to these ethical principles, we aim to contribute to the responsible
advancement of Natural Language Processing technologies and promote ethical
practices in language annotation research.

\raggedbottom
\pagebreak

\bibliography{papers}
\bibliographystyle{acl_natbib}

\clearpage
\appendix
\section{Screenshots of Various Interfaces}
\label{sec:screenshots}

\begin{figure}[!h]
    \includegraphics[width=\columnwidth]{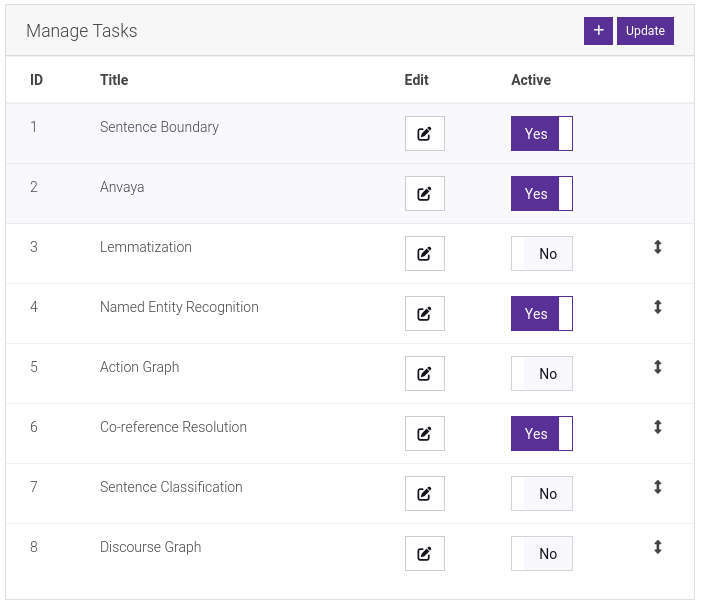}
    \caption{Task Management Interface: Add, Edit, Activate, Deactivate, Reorder Tasks}
\label{fig:interface_admin_manage_tasks}
\end{figure}

\begin{figure}[!h]
    \includegraphics[width=\columnwidth]{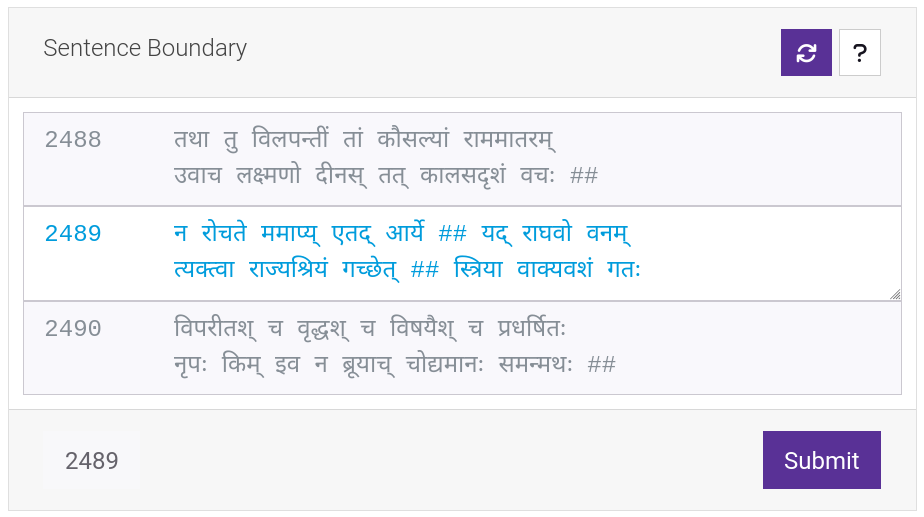}
    \caption{Sentence Boundary Annotation Interface}
    \label{fig:interface_boundary}
\end{figure}

\begin{figure}[!h]
    \includegraphics[width=\columnwidth]{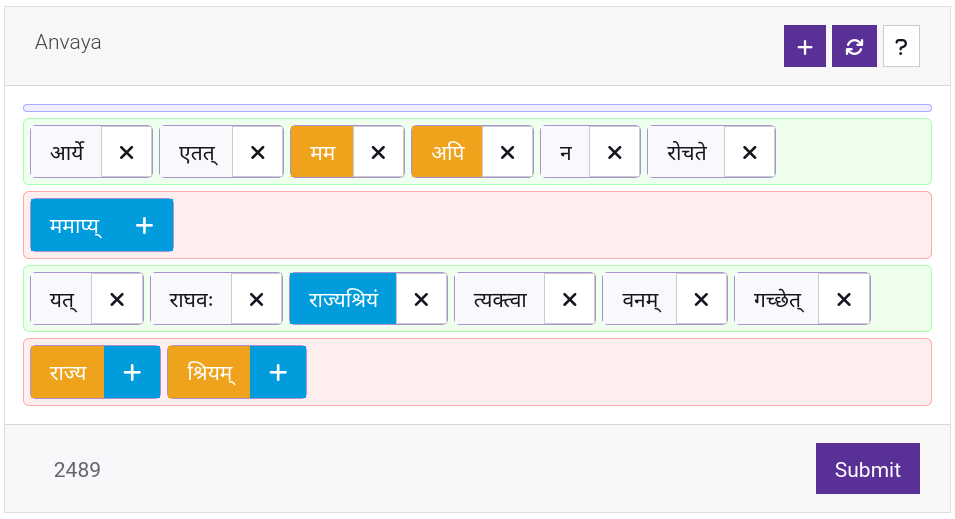}
    \caption{Word Order Annotation Interface}
    \label{fig:interface_word_order}
\end{figure}

\begin{figure}[t]
    \centering
    \includegraphics[width=0.7\columnwidth]{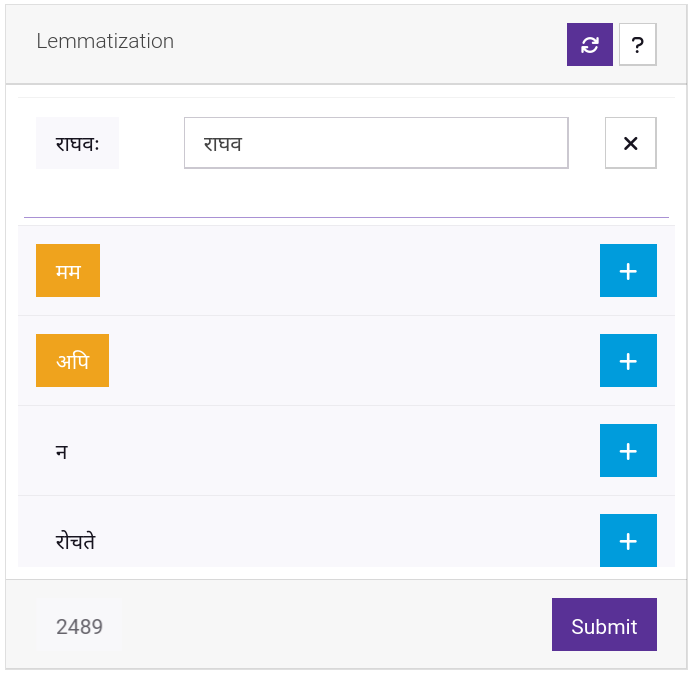}
    \caption{Token Annotation Interface: Lemmatization}
    \label{fig:interface_token_annotation}
\end{figure}

\begin{figure}[!h]
    \centering
    \includegraphics[width=0.8\columnwidth]{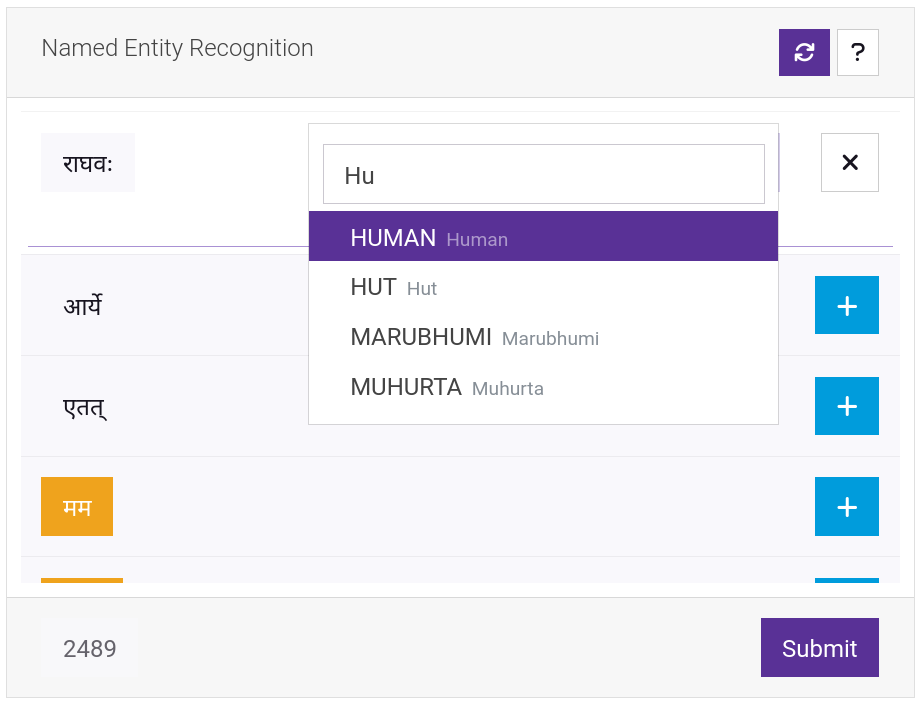}
    \caption{Token Classification Interface: Named Entity Recognition}
    \label{fig:interface_token_classification}
\end{figure}

\begin{figure}[!h]
    \centering
    \includegraphics[width=\columnwidth]{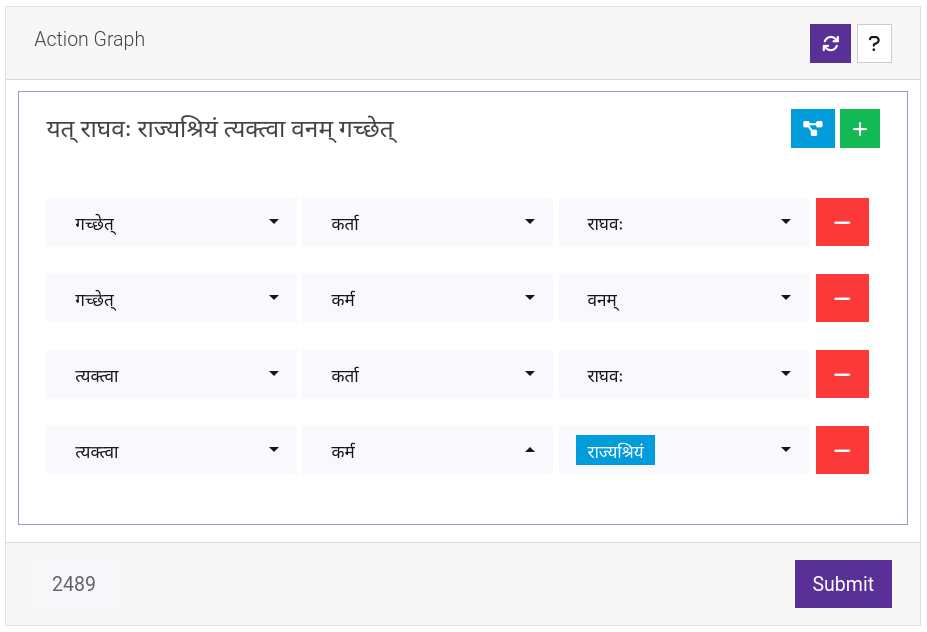}
    \includegraphics[width=0.9\columnwidth]{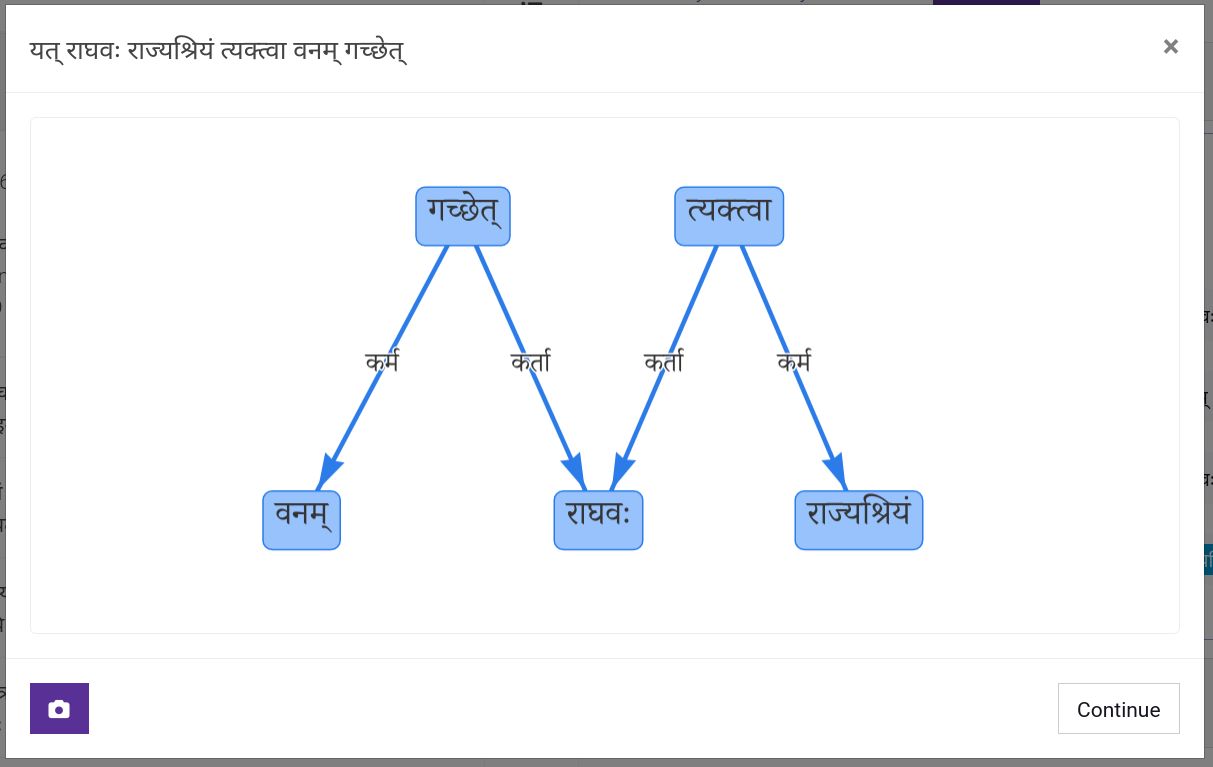}
    \caption{Token Graph Interface with Graph Visualization: Action Graph}
\label{fig:interface_token_graph}
\end{figure}

\begin{figure}[!h]
    \includegraphics[width=\columnwidth]{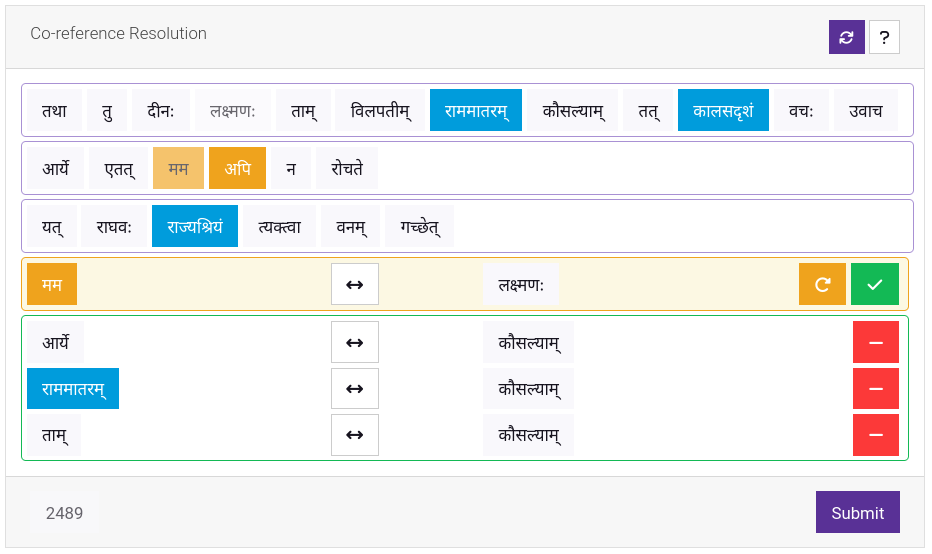}
    \caption{Token Connection Interface: Co-reference Resolution}
\label{fig:interface_token_connection}
\end{figure}

\begin{figure}
    \includegraphics[width=0.9\columnwidth]{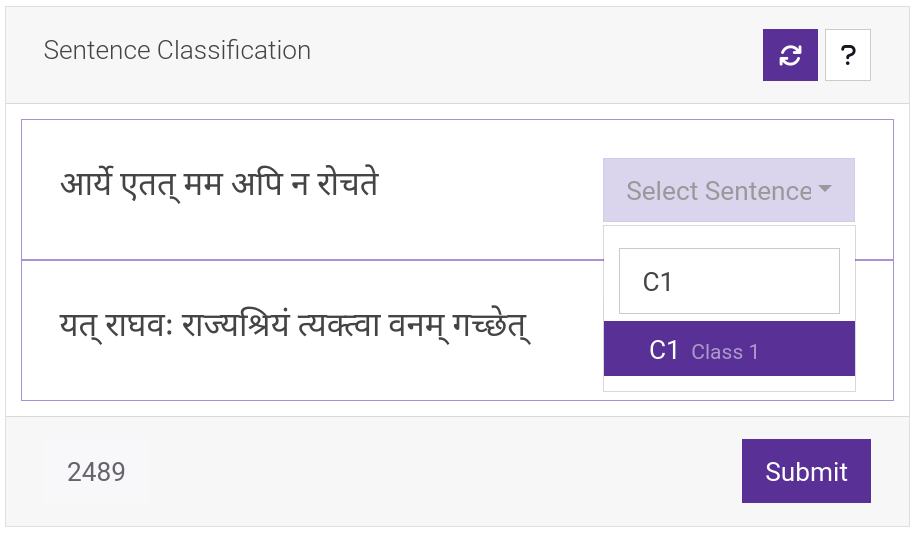}
    \caption{Sentence Classification Interface}
    \label{fig:interface_sentence_classification}
\end{figure}

\begin{figure}[!h]
    \centering
    \includegraphics[width=\columnwidth]{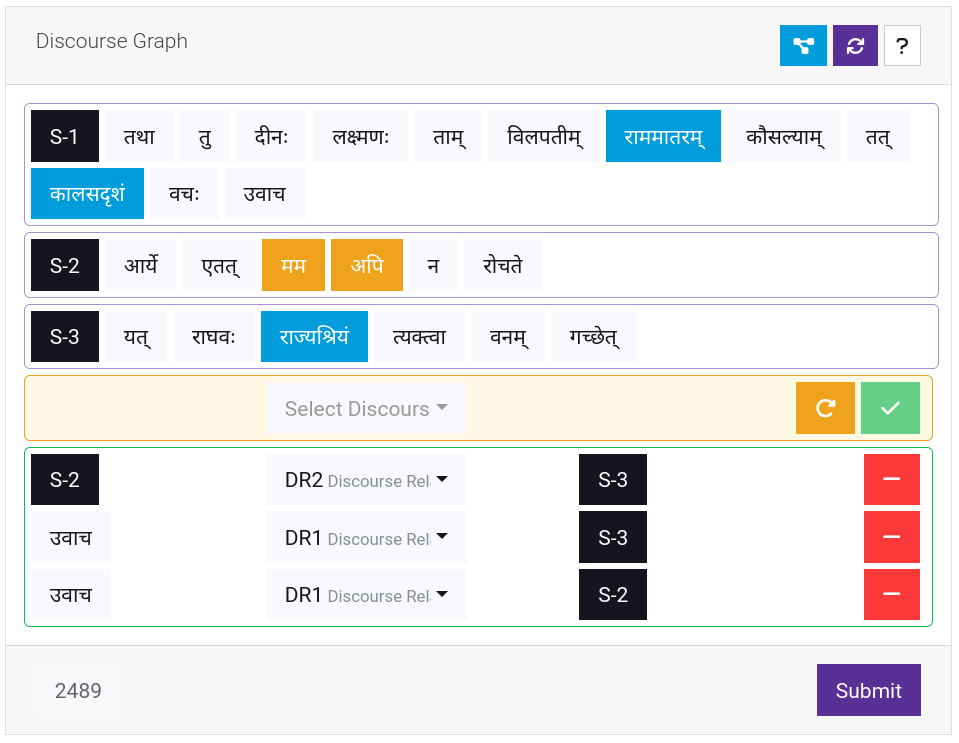}
    \caption{Sentence Graph Interface}
    \label{fig:interface_sentence_graph}
\end{figure}

\begin{figure*}
    \includegraphics[width=\textwidth]{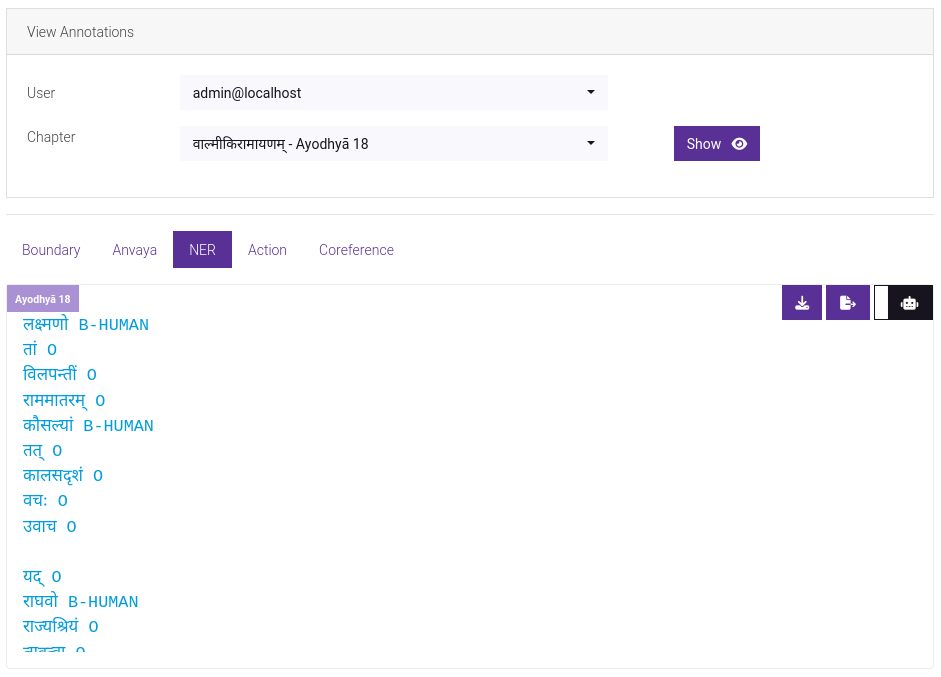}
    \caption{Export Interface: NER data in the standard BIO format}
    \label{fig:interface_export}
\end{figure*}

We showcase various interfaces of \emph{\name} in this section. An
administrative interface for managing tasks is shown in
\cref{fig:interface_admin_manage_tasks}.
\Cref{fig:interface_boundary,fig:interface_word_order,fig:interface_token_annotation,fig:interface_token_classification,fig:interface_token_graph,fig:interface_token_connection,fig:interface_sentence_classification,fig:interface_sentence_graph}
illustrate annotation interfaces for each task category.
\cref{fig:interface_export} highlights the export interface with the capability
to export the data in the standard format.

\section{Schema}
\label{sec:schema}

\begin{figure*}[b]
    \centering
    \includegraphics[width=0.8\textwidth]{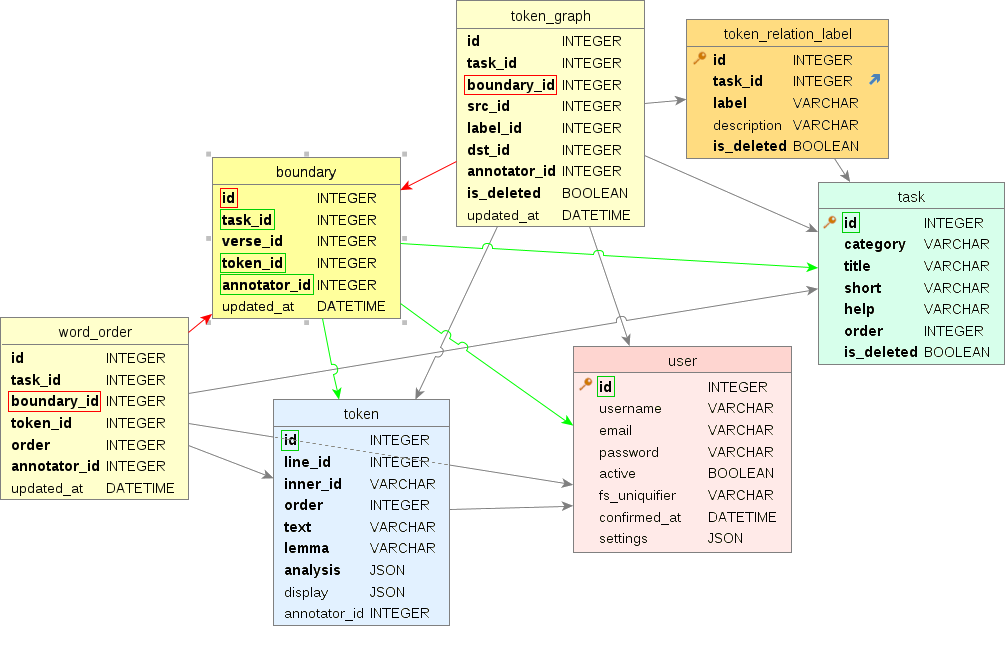}
    \vspace*{-2mm}
    \caption{Entity Relationship Diagram illustrating some relevant links.
    Tables are color coded. Yellow: Annotation Tables, Orange: Ontology Tables,
    Blue: Corpus Tables, Pink: User Tables, Green: Task Information Table. The
    annotation table for `Sentence Boundary' task is highlighted, showing the
    references incoming (red) and outgoing (green) references to other tables.}
\label{fig:entity_relation_diagram}
\end{figure*}

\emph{\name} utilizes a relational database to store information such as, corpus
data, user data, task data and annotations. A relational database allows for
efficient storage and retrieval of functional data, as well as the ability to
establish relationships between different pieces of data. For example,
annotations of specific verses by specific users can be linked allowing the
system to quickly locate and display relevant annotations when needed.

\subsection{Tasks}

The information regarding tasks is stored in a single table within the
relational database. This table serves as a centralized repository for
information related to each task, including its title, category, and
instructions for annotators. Each task is assigned a unique identifier known as
a `task id', which serves as a means of easily referring or linking to a
specific task.

\subsection{Ontology}

Ontology is required for four task categories: token classification, token
graph, sentence classification, and sentence graph. The ontology information is
stored as a flat list of labels in four separate tables, each specific to a
particular task category. There may be multiple tasks corresponding to each of
these categories. Therefore, every ontology table also has a `task id' column
which associates the ontology entries with the corresponding tasks. This setup
allows for clear organization and linking of the ontology information with the
relevant tasks.

\subsection{Annotations}

There are eight annotation tables, each corresponding to a different category of
annotation tasks. Annotations of all tasks belonging to each category are stored
in the corresponding table. The annotations are linked to the semantic units of
text, specifically, the sentences marked in the sentence boundary task. The
other seven annotation tables include a reference to the `boundary id'. In cases
where the sentence boundary task is not necessary, the boundaries of the
annotation text units (e.g., verse) are considered as sentence boundaries and
annotated automatically in the background using a special annotation user.
Additionally, to facilitate multiple instances of tasks from each task category,
every annotation table contains a reference to the `task id'. Finally, each
annotation table sports a tailored schema to support the recording of task
specific annotations. An `annotator id' associated with every task annotation
table, allows for proper organization and tracking of the annotations.

\cref{fig:entity_relation_diagram} shows the Entity Relationship (ER) diagram on
a subset of tables from the relational database of \emph{\name}.
  
\end{document}